\pdfoutput=1

\documentclass[11pt]{article}

\usepackage{multirow}
\usepackage{subcaption}
\usepackage{graphicx}
\usepackage{amsmath}
\usepackage{amssymb}

\usepackage{ACL2023}

\usepackage{times}
\usepackage{latexsym}

\usepackage[T1]{fontenc}

\usepackage[utf8]{inputenc}

\usepackage{microtype}

\usepackage{inconsolata}

\title{TreeSeg: Hierarchical Topic Segmentation of Large Transcripts}

\author{
    Dimitrios C. Gklezakos\hspace{1em} Timothy Misiak\hspace{1em} Diamond Bishop \\
    Augmend \\
    \texttt{\{dimi, tim, diamond\}@augmend.com}
}

\begin{document}
\maketitle
\begin{abstract}
From organizing recorded videos and meetings into chapters, to breaking down large inputs in order to fit them into the context window of commoditized Large Language Models (LLMs), topic segmentation of large transcripts emerges as a task of increasing significance. Still, accurate segmentation presents many challenges, including (a) the noisy nature of the Automatic Speech Recognition (ASR) software typically used to obtain the transcripts, (b) the lack of diverse labeled data and (c) the difficulty in pin-pointing the ground-truth number of segments. In this work we present TreeSeg, an approach that combines off-the-shelf embedding models with divisive clustering, to generate hierarchical, structured segmentations of transcripts in the form of binary trees. Our approach is robust to noise and can handle large transcripts efficiently. We evaluate TreeSeg on the ICSI and AMI corpora, demonstrating that it outperforms all baselines. Finally, we introduce TinyRec, a small-scale corpus of manually annotated transcripts, obtained from self-recorded video sessions.
\end{abstract}

\section{Introduction}
The wide availability of video conferencing platforms, together with the rapid surge in the volume of hosted videos \cite{McGrady_Zheng_Curran_Baumgartner_Zuckerman_2023}, have resulted in the proliferation of self-recorded content in the form of meetings and videos. Often transcribed into text via Automatic Speech Recognition (ASR), this content offers a wealth of information waiting to be extracted.

In this work we focus on segmenting large transcripts originating from automatically transcribed, self-recorded content, into temporally contiguous but semantically distinct segments. The goal of segmentation in this context is two-fold: (a) To display content in an organized manner (i.e. automatic chapter generation) and (b) to break down large transcripts in order to satisfy the size constraints of downstream models, such as the context window limitations of commoditized Large Language Models (LLMs).

In this context, topic segmentation poses many challenges due to: (a) The noisy nature of ASR software, resulting in errors due to poor transcription or out-of-dictionary technical terms, (b) the scarcity of labeled data of a diverse distribution and the difficulty in obtaining it \citet{inbook_anno_diff} and (c) the difficulty in pin-pointing the ground-truth number of segments, which can vary between human annotators even for the same transcript.

In this paper we propose TreeSeg, a novel hierarchical topic segmentation approach. TreeSeg combines utterance embeddings with divisive clustering to filter the input and identify segment transition points. Our approach is completely unsupervised, has no learnable parts and utilizes readily available off-the-shelf embedding models. TreeSeg partitions the input in a hierarchical manner and is accurate at multiple levels of segmentation resolution. In the context of automatically generating and displaying video or meeting segmentations, this hierarchical aspect of TreeSeg  provides the user with the affordance to dynamically choose the desired number and resolution of the generated chapters/segments. 

We evaluate our approach on two standard large meeting corpora: ICSI \citep{1198793} and AMI \citep{ami_article}. We demonstrate that TreeSeg outperforms its competition across the board. We also contribute a small-scale corpus of our own, TinyRec, consisting of $21$ self-recorded sessions with technical content, transcribed via ASR and manually annotated. We plan to gradually extend TinyRec over time with more annotated sessions.

\subsection{Related Work}
\citet{Koshorek2018TextSA} adopt a supervised learning approach to topic segmentation of written text by applying an LSTM-based \citep{hochreiter1997long} model to WIKI-727K, a dataset extracted from Wikipedia. Models based on transformers \citep{10.5555/3295222.3295349} such as BERT \citep{Devlin2019BERTPO} and its variants \citep{DBLP:journals/corr/abs-1907-11692} are considered by \citet{lukasik-etal-2020-text} and \citet{10.1007/978-3-031-47718-8_7}. \citet{Retkowski2024FromTS} use annotations obtained from Youtube videos to train a transformer-based segmentation model in a supervised manner.

\citet{bayomi-lawless-2018-c} apply agglomerative clustering to extract a hierarchy of segments from text. \citet{hazem-etal-2020-hierarchical} use a bottom-up approach to segment text from medieval manuscripts. \citet{grootendorst2022bertopic} uses Sentence-BERT \citep{reimers-gurevych-2019-sentence} embeddings  to cluster documents and extract latent topics.

Unsupervised topic segmentation of meetings has generated a lot of interest in the recent years. Most recent approaches are essentially modern variants of TextTiling \citep{10.5555/972684.972687}, a technique that relies on similarities between adjacent utterances. TextTiling identifies topic changes by finding local similarity minima. Perhaps the closest work to our own is BertSeg, introduced in \citet{Solbiati2021UnsupervisedTS}. BertSeg is a modern version of TextTiling that embeds utterances using a pre-trained model, extracts overlapping blocks of embeddings and aggregates them to compute utterance similarities. HyperSeg, introduced in \citet{Park_2023}, also follows the TextTiling paradigm, while replacing learned embeddings with hyper-dimensional vectors derived from random word embeddings. CohereSeg \citep{xing-carenini-2021-improving} and $\text{M}^3$Seg \citep{wang-etal-2023-m3seg} are unsupervised segmentation approaches that fine-tune embeddings using what is, in essence, a contrastive learning technique. CohereSeg focuses on dialogue topic segmentation and is shown in \citet{Park_2023} to perform worse than or on par with HyperSeg. At the time of writing this manuscript there is no publicly available code base for $\text{M}^3$Seg. Finally, \citet{ghosh2022topic} compare various topic segmentation approaches on semi-structured and unstructured conversations and show that pre-training on structured data does not transfer well to unstructured data.

\section{Method}
Consider the linear topic segmentation setting, where the input is a temporal sequence of transcript entries/utterances $U = [U_1,...,U_T]$. We will henceforth refer to $U$ as the `timeline'. The underlying organization of the transcript into topics is modeled as a partition $P=\{P_k\}_{k=1}^{K}$ of $U$ into segments. Each such segment covers a temporally contiguous set of utterances:
$$
P_k=\{U_t : t_s(k)\leq t \leq t_e(k)\}
$$
starting with $U_{t_s(k)}$ and ending with $U_{t_e(k)}$ (endpoints included). Each utterance belongs to exactly one segment. The goal of linear topic segmentation is to approximate this ground-truth partition.

\subsection{Hierarchical Topic Segmentation}

We extend the linear setting to a hierarchical version by considering nested partitions of the timeline, of increasing resolution, represented by trees. 

\begin{figure}[h!]
\centering
    \begin{subfigure}{0.47\textwidth}
    \centering
    \includegraphics[width=0.9\linewidth]{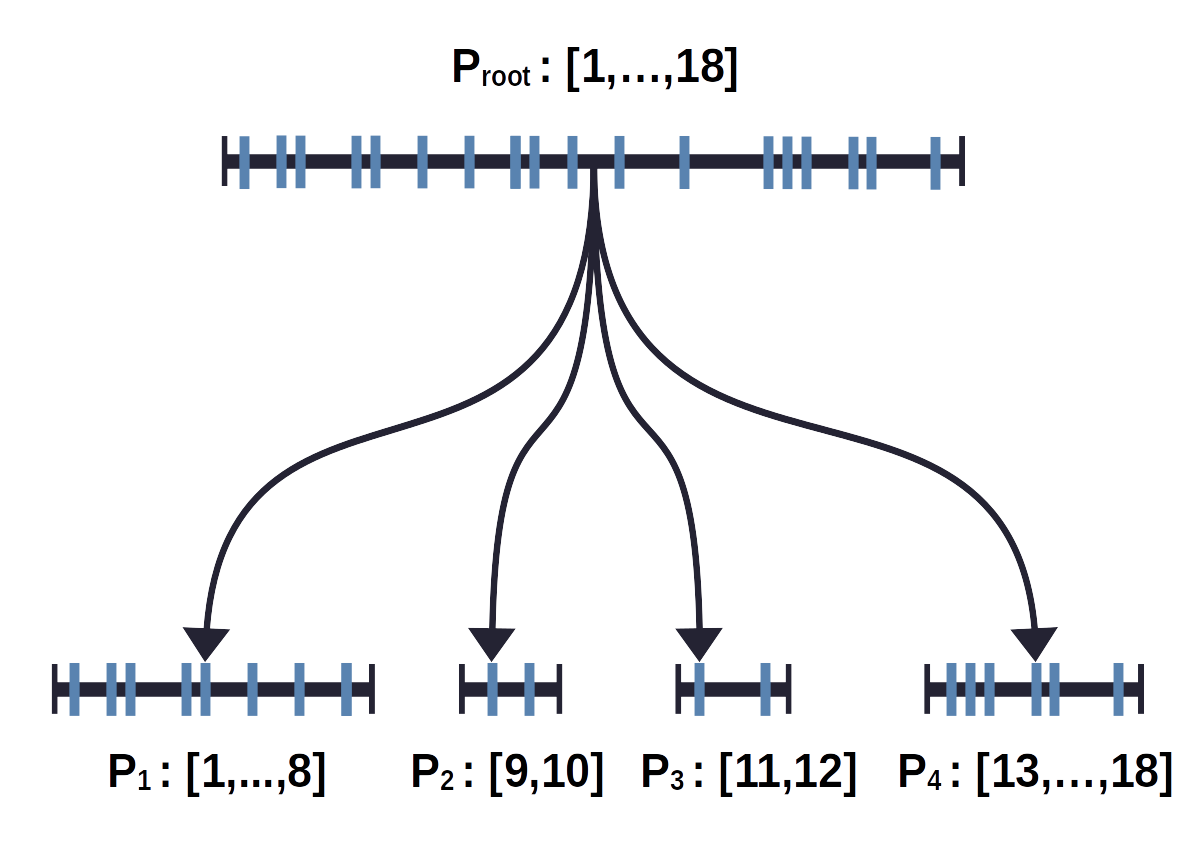}
    \caption{\textbf{Flat partition tree:} Linear topic segmentation.}
    \label{fig:partition:flat}
    \end{subfigure}
    \begin{subfigure}{0.47\textwidth}
    \vspace{5ex}
    \centering
    \includegraphics[width=0.9\linewidth]{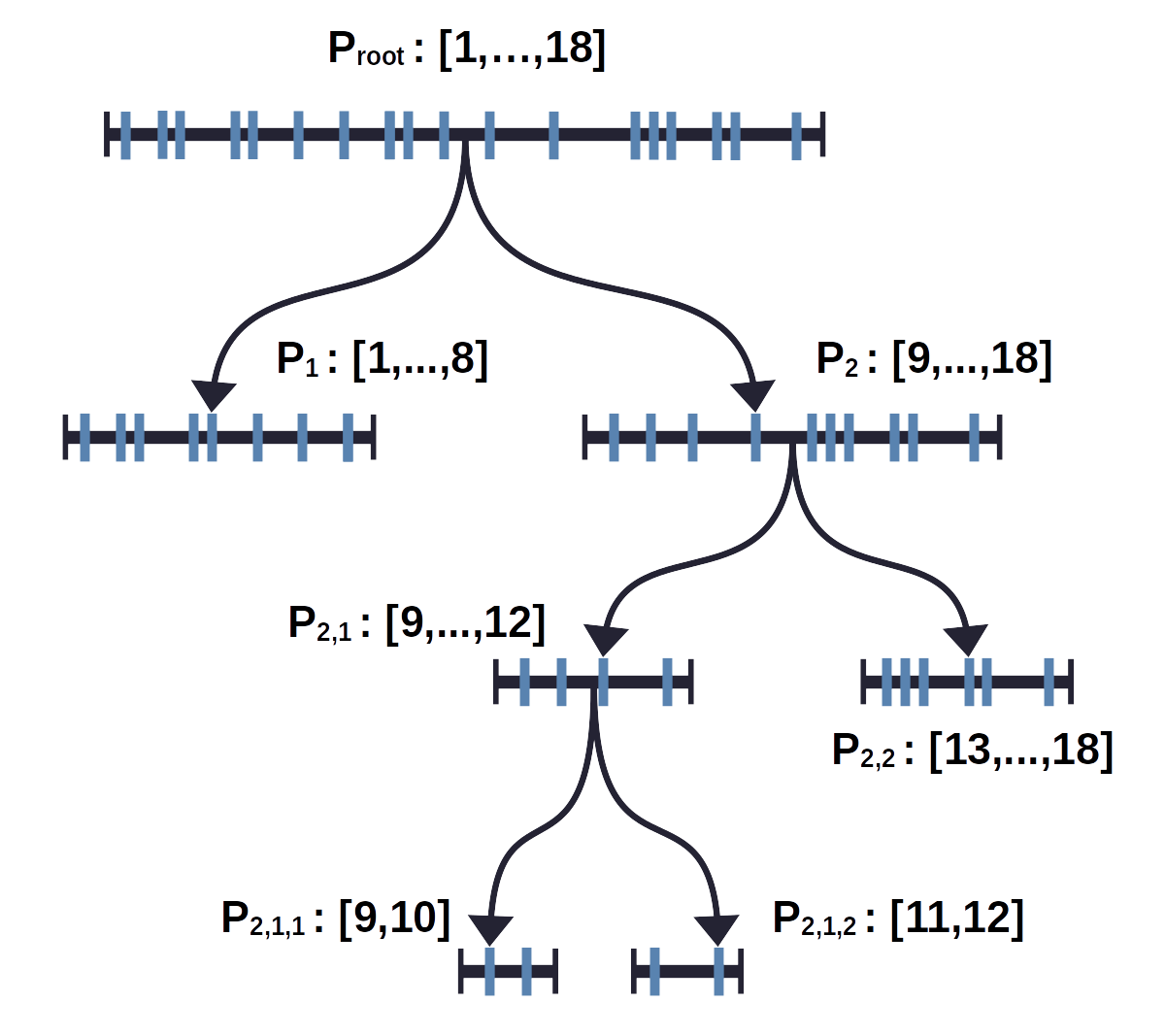}
    \caption{\textbf{Deep partition tree:} Hierarchical topic segmentation.}
    \label{fig:partition:full}
    \end{subfigure}
    \caption{\textbf{From linear to hierarchical topic segmentation:} \textbf{(a)} A partition tree of depth equal to $1$ corresponding to linear topic segmentation and \textbf{(b)} a deeper partition tree corresponding to hierarchical topic segmentation. The root node always covers the full timeline. Note that in both cases, the children of a node form a partition of the node's segment.}
    \label{fig:partition}
\end{figure}

A flat partition $P$ of the timeline can be viewed as a tree of depth equal to $1$, where the root is the timeline itself and each child is an utterance set $P_k$ in the partition (see Figure \ref{fig:partition:flat}). A nested partition is represented by a deeper tree where each node $v$ corresponds to a temporally contiguous set of utterances $P_v$, while its children, denoted by $N(v)$ form a partition of $P_v$, such that: $P_v = \bigcup_{u\in N(v)}P_u$ and $\forall i,j \in N(v)\; :\; i\neq j$, $P_{i}\cap P_{j} = \varnothing$.

Figure \ref{fig:partition:full} shows an example of such a nested partition and its corresponding tree. A partition tree has the following properties:
\begin{enumerate}
    \item Every sub-tree containing the root is a valid nested partition of the timeline.
    \item The leaves of every sub-tree containing the root form a valid flat partition of the timeline.
\end{enumerate}
The partition tree also induces a natural order in which segments are divided into sub-segments. Consider the sub-tree that contains all nodes of depth smaller than $\tau\leq D(P)$, where $D(P)$ is the maximum depth of the partition tree and let $P_{\leq\tau}$ denote its corresponding valid, nested partition. The leaves of this sub-tree form a flat partition of the timeline. As $\tau$ increases, the resolution $|P_{\leq\tau}|$ of this flat partition increases as well.

\subsection{From Linear to Hierarchical Partitions}
\label{sec:lin-hier}
Suppose that we have access to a linear topic segmentation model that takes in the desired partition length $K$ and identifies $K$ segments. Suppose also that as $K$ increases, additional segment boundaries are added but not deleted\footnote{Note that techniques based on TextTiling \citep{10.5555/972684.972687} such as BertSeg \citep{Solbiati2021UnsupervisedTS} and HyperSeg \citep{Park_2023} naturally exhibit this property.}.  Then we can evaluate this model on the hierarchical segmentation task, as follows:
\begin{itemize}
    \item Choose a depth threshold $\tau$ and construct the corresponding partition $P_{\leq\tau}$ of maximum depth $\tau$.
    \item Query the model with $K=|P_{\leq\tau}|$.
    \item Evaluate the result by comparing with the flat partition induced by the leaves of $P_{\leq\tau}$.
    \item Repeat for all $1\leq \tau \leq D(P)$.
\end{itemize}

\begin{figure}[h!]
\centering
    \begin{subfigure}{0.45\textwidth}
    \centering
    \includegraphics[width=0.9\linewidth]{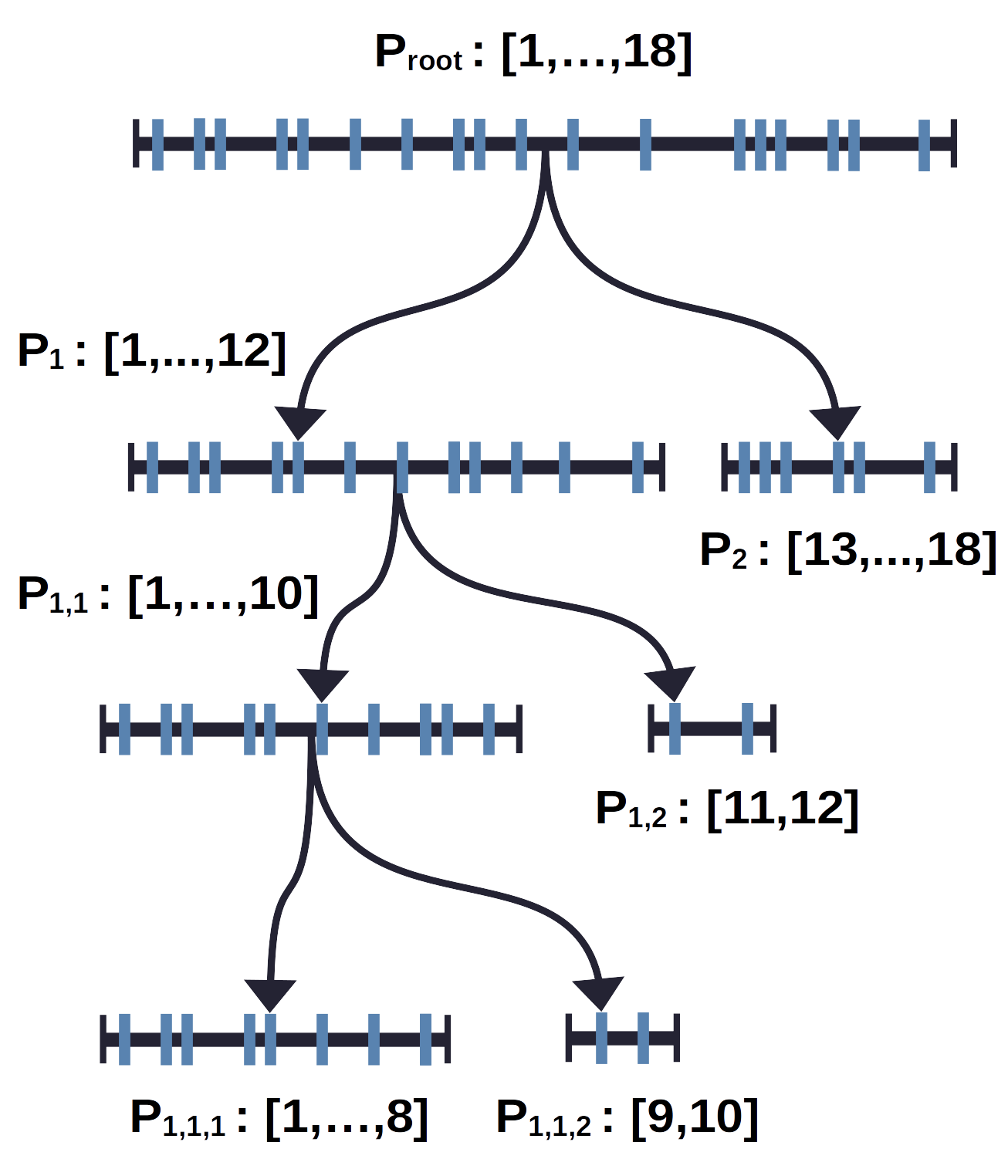}
    \end{subfigure}
   
    \caption{\textbf{Inaccurate hierarchical segmentation:} An example of an accurate linear, but inaccurate hierarchical approximation of the tree in Figure \ref{fig:partition:full}. Note that the leaves of the output partition match those of the ground-truth partition, however the order in which the nodes are partitioned is not respected and the hierarchical structure of the segments is not properly identified.}
    \label{fig:partition:wrong}
\end{figure}
Intuitively, a model that accurately captures the hierarchical relations between segments, will output intermediate partitions that match those induced by the limited-depth sub-trees of the ground truth partition. Figure \ref{fig:partition:wrong} shows an example where an output partition perfectly matches the ground-truth bottom-level partition in the linear topic segmentation setting, but fails to accurately capture its hierarchical structure.

\subsection{TreeSeg}
TreeSeg first embeds the transcript timeline entries $U$. Then these embeddings are combined with a divisive clustering approach, to identify appropriate splitting points and construct a deep partition tree.

\subsubsection{Embedding the Transcript}
TreeSeg uses an embedding model to convert transcript entries to embeddings. For the results in this paper we used OpenAI's \textit{text-embedding-ada-002} (ADA) \citep{DBLP:journals/corr/abs-2201-10005}, an embedding model trained in an unsupervised manner, utilizing contrastive learning techniques. The choice of an off-the-shelf, commoditized embedding model, results in a pipeline with no trainable parts. Note that TreeSeg does not depend on this particular choice. Any suitable embedding model such as \citet{DBLP:journals/corr/abs-1907-11692} can be directly plugged into our approach in the same manner.

A single utterance might not contain enough context to be embedded on its own in a meaningful way, especially in the presence of automatic transcription errors. To address this issue we extract overlapping blocks of utterances. For each utterance $u_t$ at position $t$ we extract a block that consists of $u_t$ itself and up to $W$ utterances in the immediate past. This block $[u_{t-W},...,u_t]$ is passed through the embedding model $f$ to obtain the block embedding $e_t = f([u_{t-W},...,u_t])$. Note that this is a point of deviation from BertSeg \citep{Solbiati2021UnsupervisedTS}. While BertSeg embeds each utterance separately and aggregates them using max-pooling, TreeSeg embeds the whole block of utterances together, resulting in additional context passed to the embedding model. Repeating for every utterance in the transcript results in a temporal sequence of utterance embeddings that maintain local context. The utterance block width $W$ is the sole hyperparameter of TreeSeg. 

\subsubsection{Divisive Clustering}
\begin{figure*}[h!]
\centering
    \begin{subfigure}{\textwidth}
    \centering
    \includegraphics[width=0.6\linewidth]{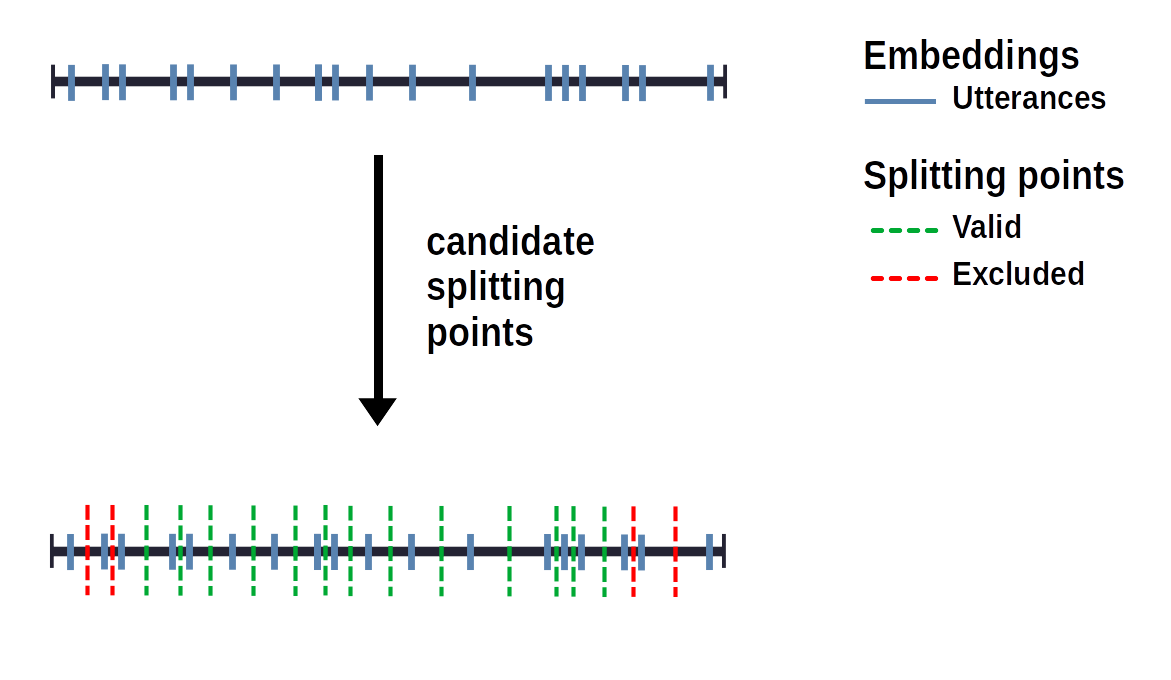}
    \caption{\textbf{Candidate splitting points:} An example utterance embedding timeline with valid candidate splitting points shown in \textit{green}. Candidates in \textit{red} are excluded due to minimum size constraints.}
    \label{fig:tree:timeline}
    \end{subfigure}
    \vspace{0.05\textwidth}
    \begin{subfigure}{0.45\textwidth}
    \centering
    \includegraphics[width=\linewidth]{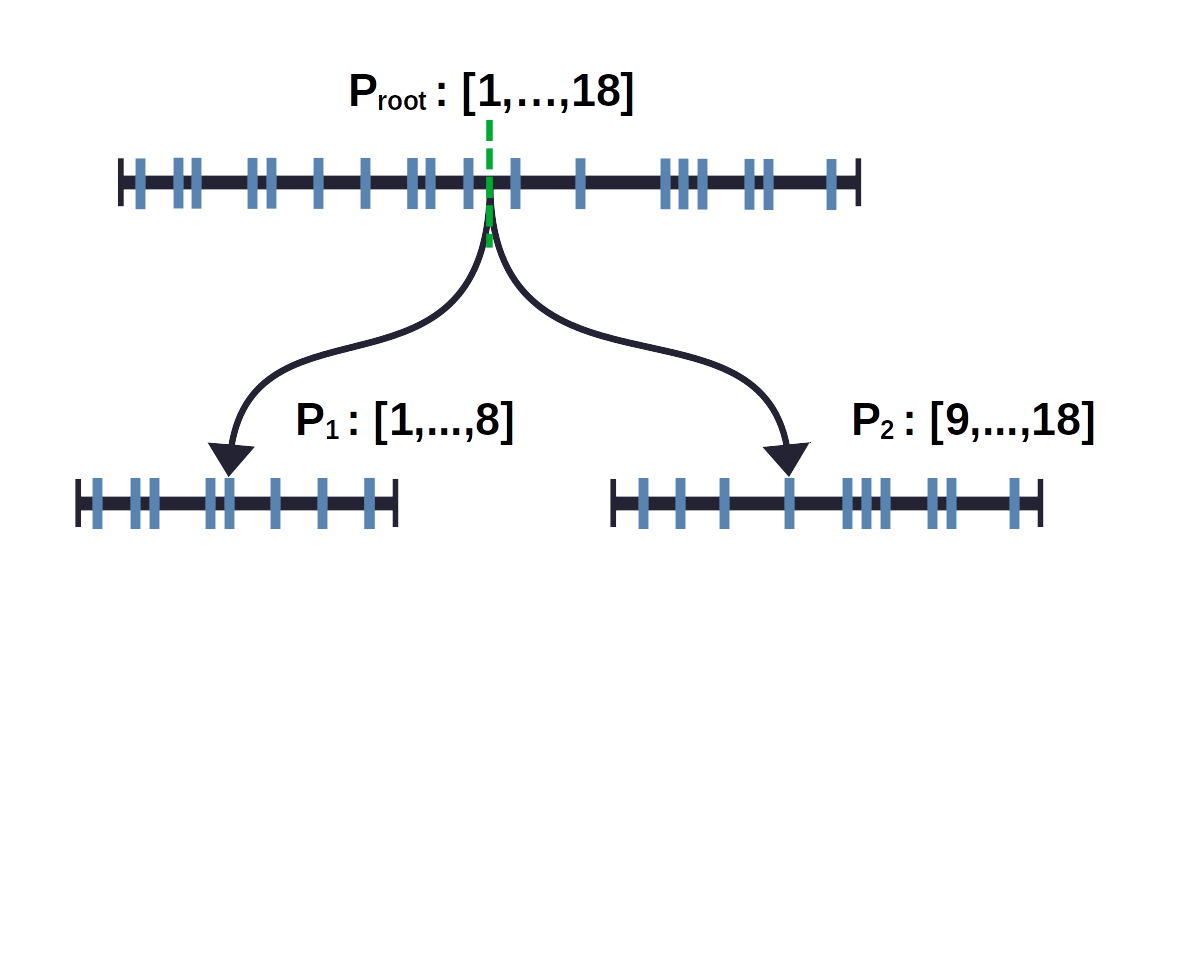}
    \caption{\textbf{First split:} The original timeline is divided into two sub-segments.}
    \label{fig:tree:split1}
    \end{subfigure}\hspace{0.05\textwidth}
    \begin{subfigure}{0.45\textwidth}
    \centering
    \includegraphics[width=\linewidth]{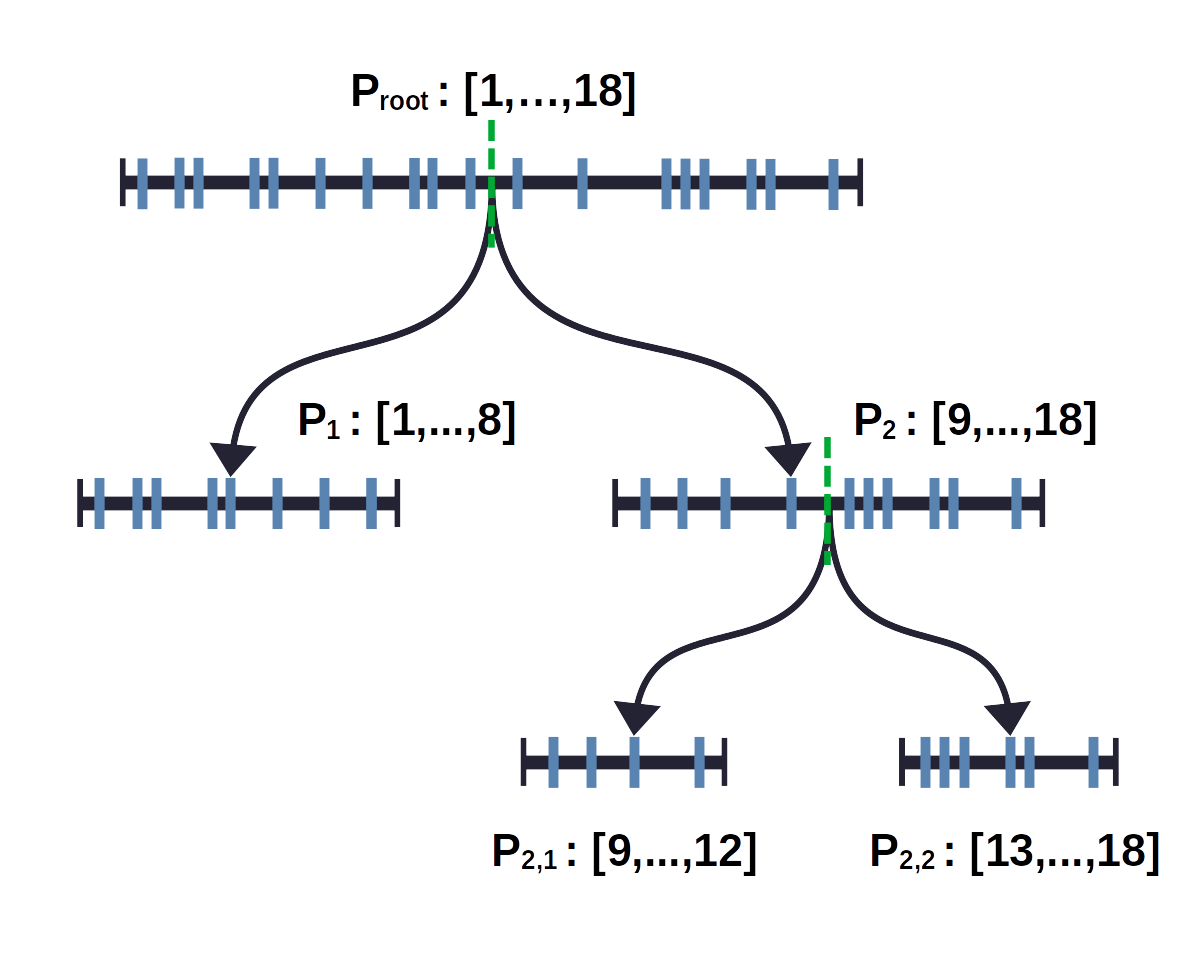}
    \caption{\textbf{Second split:} A sub-segment is sub-divided again and the partition tree deepens.}
    \label{fig:tree:split2}
    \end{subfigure}
    
\caption{\textbf{Dividing the timeline:} \textbf{(a)} At each step, valid candidate splitting points are identified for all leaves. \textbf{(b) \& (c)} The optimal splitting point across all leaves is used to divide the corresponding segment into two sub-segments. The process continues until a termination criterion is met.}
\label{fig:tree:main}
\end{figure*}

TreeSeg utilizes a divisive clustering approach to recursively split segments into two sub-segments, constructing a binary partition tree in the process. Below, we provide a high-level outline of this segment division process:

\begin{itemize}
    \item For each leaf in the current partition tree, identify the optimal splitting point according to the loss function, while respecting any constraints.
    \item Pick the leaf with the best scoring candidate splitting point and split it into two sub-segments.
    \item Repeat until the termination condition is met.
\end{itemize}

We use a one-dimensional clustering objective as our loss function. Consider a timeline of embeddings $E=[e_1,...,e_T]$. A candidate splitting point $i$ partitions this timeline into two segments $[e_1,...,e_{i-1}]$ and $[e_i,...,e_T]$. In practice we want to avoid ending up with trivial segments, therefore we enforce a minimum viable segment size denoted by $M$ and only consider candidate points in the range $M<i\leq T-M$. The loss function for candidate point $i$ is given by:
$$
\mathcal{L}(i) = \sum_{t=1}^{i-1} \| e_t - \mu_{\text{L}}\|_2^2 + \sum_{t=i}^{T} \| e_t - \mu_{\text{R}}\|_2^2
$$
with $\mu_{\text{L}} = \frac{1}{i-1}\sum_{t=1}^{i-1} e_t$ and $\mu_{\text{R}} = \frac{1}{T-i+1}\sum_{t=i}^{T} e_t$. The process stops when we reach the desired number of segments $K$ or when all leaf segments are of size $<2M$ \footnote{Further splitting such a segment will result in at least one sub-segment below the size threshold $M$.}. The division process of TreeSeg is outlined in Figure \ref{fig:tree:main}. Figure \ref{fig:tree:timeline} shows an example timeline together with the corresponding set of candidate splitting points, while Figures \ref{fig:tree:split1} and \ref{fig:tree:split2} demonstrate two successive division steps.

Note that agglomerative hierarchical clustering approaches are typically preferred to divisive ones for computational efficiency reasons. We opt for the divisive approach for the following reasons:

\begin{itemize}
\item Divisive clustering naturally matches the hierarchical segmentation task for transcripts. The termination condition is typically met long before the utterance level is reached.
\item Since timelines are one-dimensional, the optimal splitting point can be identified with a single linear pass. Efficient implementations using cumulative sums of embedding vectors allow for fast loss function computation. Optimal splits are computed only once for every node and are maintained in a min-heap data structure.
\end{itemize}
Our hypothesis is that the divisive approach utilizes global information to identify strong candidates for topic shifts, with averaging over multiple embedding vectors functioning as a candidate splitting point filter.

\section{Evaluation}
\subsection{Datasets}
We evaluate TreeSeg on three datasets:
\begin{itemize}
\item \textbf{ICSI} \citep{1198793} A corpus of $75$ transcribed meetings, containing manual hierarchical topic annotations up to $4$ levels deep.
\item \textbf{AMI} \citep{ami_article} Over $100$ hours of transcribed meetings, containing manual hierarchical topic annotations up to $3$ levels deep.
\item \textbf{TinyRec} We introduce TinyRec, a dataset consisting of transcripts obtained from $21$ self-recorded sessions. Each transcript contains spoken utterances as transcribed via ASR and was manually annotated with two-level topic annotations. For more details on the content and annotation guidelines for TinyRec, refer to Appendix \ref{app:tinyrec}
\end{itemize}
We enforce a minimum size of five utterances per segment. Segments with sizes below this threshold are automatically merged to the segment that comes immediately after them. Table \ref{tab:stats:anno} shows the topic annotation statistics for each dataset. Almost all transcripts in the ICSI and AMI corpora contain second-level annotations, while several of them contain third- or even fourth-level annotations. All TinyRec transcripts are annotated at two different resolutions: `coarse' and `fine'. Table \ref{tab:stats:seg} shows the average number of segments at each partition level, after pruning segments below the size threshold.

\begin{table}[]
\center
\begin{tabular}{lccccc}
\hline
\textbf{Dataset} & \textbf{Avg. $|U|$} & \textbf{L1} & \textbf{L2} & \textbf{L3} & \textbf{L4} \\ \hline
\textbf{ICSI}  &  1453.7 & 75          & 75          & 52          & 3           \\
\textbf{AMI}  &   636.4 & 139         & 125         & 21          & 0           \\
\textbf{TinyRec} & 267.4 & 21          & 21          & 0           & 0           \\ \hline
\end{tabular}
\caption{\textbf{Annotation statistics:} Number of transcripts with available topic annotations per partition level (\textbf{L1} through \textbf{L4}), together with the average utterance timeline length.}
\label{tab:stats:anno}
\end{table}

\begin{table}[]
\center
\begin{tabular}{lcccc}
\hline
\textbf{Dataset} & \textbf{L1} & \textbf{L2} & \textbf{L3} & \textbf{L4} \\ \hline
\textbf{ICSI}  & 5.8 & 19.05 & 28.08 & 34.33 \\
\textbf{AMI}  & 6.81 & 14.44 & 26.61 & - \\
\textbf{TinyRec}  &  4.18 & 14.12 & - & - \\
\hline
\end{tabular}
\caption{\textbf{Segment statistics:} Average number of segments per partition level (\textbf{L1} through \textbf{L4}), after pruning.}
\label{tab:stats:seg}
\end{table}

\subsection{Methodology}
We compare TreeSeg against four baselines on the hierarchical topic segmentation task. We adapt BertSeg and HyperSeg to output the top $K$ splitting points on the timeline, as described in Section \ref{sec:lin-hier}. We also compare with two naive baselines: RandomSeg and EquiSeg. RandomSeg generates $K$ random segments by picking $K-1$ segment transition points at random. EquiSeg splits the timeline into equidistant segments. 

For evaluation we use the standard $P_k$ \citep{Beeferman1999StatisticalMF} and \textit{WinDiff} \citep{pevzner-hearst-2002-critique} metrics. For each level of resolution we query each model with the ground-truth number of segments $K$ and compare the obtained partition with the ground-truth one. We average metrics across all possible partitions, as well as on a per-level basis. We run RandomSeg $100$ times and average the results.

The adapted version of HyperSeg has no hyperparameters. TreeSeg and BertSeg both use a hyperparameter that regulates the utterance embedding block width. BertSeg uses two additional hyperparameters related to utterance similarity score smoothing. The first five transcripts from each dataset were denoted as the `development' set and were used to determine reasonable values for these hyperparameters.

\subsection{Results}

\begin{table*}[htbp]
\center
\begin{tabular}{lcccccc}
\hline
\multicolumn{1}{c}{} & \multicolumn{2}{c}{\textbf{ICSI}} & \multicolumn{2}{c}{\textbf{AMI}} & \multicolumn{2}{c}{\textbf{TinyRec}} \\ \hline
\textbf{Method}      & \textbf{Pk}     & \textbf{Wd}     & \textbf{Pk}     & \textbf{Wd}    & \textbf{Pk}       & \textbf{Wd}      \\ \hline
\textbf{RandomSeg}   & 0.464 & 0.503 & 0.464 & 0.503 & 0.465 & 0.492 \\
\textbf{EquiSeg}     & 0.482 & 0.508 & 0.478 & 0.506 & 0.505 & 0.513 \\
\textbf{HyperSeg}    & 0.453 & 0.499 & 0.48  & 0.519 & 0.485  & 0.515  \\
\textbf{BertSeg}     & 0.388 & 0.432 & 0.443 & 0.48  & 0.473 & 0.486 \\
\textbf{TreeSeg (ours)}  & \textbf{0.31}  & \textbf{0.353} & \textbf{0.355} & \textbf{0.396} & \textbf{0.367} & \textbf{0.382} \\ \hline
\end{tabular}

\caption{\textbf{Hierarchical topic segmentation results:} The performance of all techniques is evaluated on the ICSI, AMI and TinyRec datasets. $P_k$ (\textbf{Pk}) and \textit{WinDiff} (\textbf{Wd}) metrics (lower is better) are aggregated over all segmentation resolution levels. Our approach (TreeSeg) clearly outperforms all baselines.
}
\label{tab:results:agg}
\end{table*}

\begin{table*}[h!]
\center
\begin{subtable}{0.9\textwidth}
\centering
\begin{tabular}{lccccccccc}
\hline
\multicolumn{1}{c}{\textbf{}} & \multicolumn{4}{c}{\textbf{ICSI (Pk)}}                     & \multicolumn{3}{c}{\textbf{AMI (Pk)}}        & \multicolumn{2}{c}{\textbf{TinyRec (Pk)}} \\ \hline
\textbf{Method}               & \textbf{L1} & \textbf{L2} & \textbf{L3} & \textbf{L4} & \textbf{L1} & \textbf{L2} & \textbf{L3} & \textbf{L1}       & \textbf{L2}      \\ \hline
\textbf{RandomSeg}  & 0.445 & 0.472 & 0.48 & 0.487 & 0.455 & 0.471 & 0.486 & 0.44 & 0.49        \\
\textbf{EquiSeg}    & 0.446 & 0.5 & 0.505 & 0.492 & 0.464 & 0.492 & 0.492 & 0.492 & 0.518        \\
\textbf{HyperSeg}   & 0.442 & 0.456 & 0.465 & 0.417 & 0.47 & 0.491 & 0.481 & 0.48 & 0.49        \\
\textbf{BertSeg}    & 0.343 & 0.416 & 0.41 & 0.422 & 0.441 & 0.442 & 0.461 & 0.462 & 0.484        \\
\textbf{TreeSeg (ours)} & \textbf{0.28} & \textbf{0.325} & \textbf{0.326} & \textbf{0.392} & \textbf{0.35} & \textbf{0.356} & \textbf{0.38} & \textbf{0.336} & \textbf{0.399}        \\ \hline
\end{tabular}
\caption{{$P_k$}}
\label{tab:results:lvl:pk}
\end{subtable}

\begin{subtable}{0.9\textwidth}
\centering
\begin{tabular}{lccccccccc}
\hline
\multicolumn{1}{c}{\textbf{}} & \multicolumn{4}{c}{\textbf{ICSI (Wd)}}                     & \multicolumn{3}{c}{\textbf{AMI (Wd)}}        & \multicolumn{2}{c}{\textbf{TinyRec (Wd)}} \\ \hline
\textbf{Method}               & \textbf{L1} & \textbf{L2} & \textbf{L3} & \textbf{L4} & \textbf{L1} & \textbf{L2} & \textbf{L3} & \textbf{L1}       & \textbf{L2}      \\ \hline
\textbf{RandomSeg}  & 0.474 & 0.516 & 0.527 & 0.534 & 0.489 & 0.515 & 0.532 & 0.461 & 0.523 \\
\textbf{EquiSeg}    & 0.47  & 0.531 & 0.53 & 0.519 & 0.493 & 0.52 & 0.515 & 0.502 & 0.524 \\
\textbf{HyperSeg}   & 0.473 & 0.508 & 0.523 & 0.481 & 0.504 & 0.534 & 0.533 & 0.506 & 0.524 \\
\textbf{BertSeg}    & 0.386 & 0.462 & 0.453 & 0.466 & 0.477 & 0.481 & 0.498 & 0.479 & 0.493 \\
\textbf{TreeSeg (ours)} & \textbf{0.314} & \textbf{0.375} & \textbf{0.372} & \textbf{0.441} & \textbf{0.387} & \textbf{0.403} & \textbf{0.421} & \textbf{0.352} & \textbf{0.413} \\ \hline
\end{tabular}
\caption{\textit{WinDiff}}
\label{tab:results:lvl:wd}
\end{subtable}
\caption{
\textbf{Hierarchical topic segmentation results aggregated per level:} The performance of all techniques is evaluated on the ICSI, AMI and TinyRec datasets. \textbf{(a)} \textit{WinDiff} (\textbf{Wd}) and \textbf{(b)} $P_k$ (\textbf{Pk}) metrics are aggregated per segmentation resolution level. Our approach (TreeSeg) maintains strong performance across all segmentation resolutions.
}
\label{tab:results:lvl}
\end{table*}

Table \ref{tab:results:agg} shows the $P_k$ and \textit{WinDiff} scores for all approaches, averaged over all topic annotation resolutions as described in Section \ref{sec:lin-hier}. TreeSeg clearly outperforms all baselines on all three datasets. Table \ref{tab:results:lvl} shows the $P_k$ (Table \ref{tab:results:lvl:pk}) and \textit{WinDiff} (Table \ref{tab:results:lvl:wd}) scores of all approaches aggregated per segmentation resolution level. Note that TreeSeg maintains strong performance across all segmentation resolutions.

Our results demonstrate that TreeSeg adequately captures the hierarchical relations between segments at all levels of the hierarchy. Note also that the performance of BertSeg and HyperSeg degrades on TinyRec, a dataset that is less structured in nature than ICSI or AMI. While small in scale, TinyRec might be more representative of self-recorded content in the wild. TreeSeg maintains strong performance across all three datasets.

\section{Conclusion}
We introduced TreeSeg, a hierarchical segmentation approach suitable for segmenting large meeting and video transcripts. TreeSeg generates structured segmentations in the form of binary trees, capturing the hierarchical relations between segments. Our approach utilizes off-the-shelf components, contains no learnable parts and only a single hyperparameter. We provided a rigorous definition and evaluation methodology for the hierarchical topic segmentation task. We compared TreeSeg with two related embeddings-based approaches ; BertSeg \citep{Solbiati2021UnsupervisedTS} and HyperSeg \citep{Park_2023}, as well as two naive baselines. We introduced TinyRec, a small-scale collection of transcripts obtained from self-recorded sessions via ASR. Evaluating on ICSI, AMI and TinyRec, we demonstrated the superior performance of TreeSeg. Our work constitutes, to our knowledge, the first divisive clustering variant of TextTiling \citep{10.5555/972684.972687}. A promising future direction for research is that of utilizing the structure of the generated partition in downstream tasks such as summarization \citep{Park_2024} or knowledge extraction.

\section{Limitations}
One limitation of this work is related to the diversity of the datasets in our evaluation. ICSI and AMI are large corpora but are unlikely to capture the full diversity of self-recorded content. Contributing TinyRec is one attempt at mitigating this limitation.
Another limitation is the lack of comparison with $\text{M}^3$Seg  \citep{wang-etal-2023-m3seg}. At the time of writing this manuscript, there is no publicly available code base for $\text{M}^3$Seg. Evaluation metrics on ICSI and AMI vary with the resolution of the segmentation and the data extraction process, making evaluating models on the same exact data a necessity for a fair comparison. Finally another limitation of this work is our restricted focus on ADA embeddings \citep{DBLP:journals/corr/abs-2201-10005}. A more thorough comparison of various embedding models might yield interesting insights into the function of divisive clustering as a filter for strong topic shift candidate points.

\bibliographystyle{acl_natbib}
\bibliography{trees}

\appendix

\section{The TinyRec Dataset}
\label{app:tinyrec}
TinyRec consists of $21$ self-recorded video sessions with screen-sharing that were transcribed using Automatic Speech Recognition (ASR) and manually annotated with topic annotations at two levels of resolution. The selection criteria were the following:
\begin{itemize}
\item A session needs to be at least $8$ minutes long or contain at least $80$ transcript entries.
\item The content of the session must be technical in nature and non-trivial.
\end{itemize}

The dataset was annotated by four different annotators. The concept of a partition was explained to each annotator. The annotators were asked to first identify a `coarse' partition typically consisting of $2$-$5$ segments for a ten-minute session. Then they were asked to further partition each `coarse' segment, wherever that made sense according to their judgement, to obtain the `fine', two-level partition.

The transition from `coarse' to `fine' segments was explained with the following example scenario:\\

\emph{You are recording your work update for a week during which you worked on three features. Suppose that you identified a coarse segment covering the discussion on one of these features. The fine segmentation would segment this coarse segment again into sub-segments, each discussing parts of the feature implementation or nuances in its design.}
\\

as well as an example of a segmented transcript.

\end{document}